\documentclass[runningheads]{llncs}
\usepackage[T1]{fontenc}
\usepackage{graphicx}
\usepackage{color}
\graphicspath{ {./images/} }

\usepackage{caption}
\usepackage{subcaption}

\usepackage{amssymb}
\usepackage{amsmath}

\usepackage{tikz}
\usetikzlibrary{positioning}
\usetikzlibrary{shapes}
\usetikzlibrary{arrows}
\usepackage{bbm}

\newcommand\numberthis{\addtocounter{equation}{1}\tag{\theequation}}

\begin{document}
\title{Error Analysis of Shapley Value-Based Model Explanations: An Informative Perspective}
%
%
\author{Ningsheng Zhao\inst{1}
\and
Jia Yuan Yu\inst{1} 
\and
Krzysztof Dzieciolowski\inst{1,2} 
\and
Trang Bui\inst{3} 
}
\authorrunning{N. Zhao et al.}
\titlerunning{Error Analysis of Shapley Value-Based Explanations}
%
\institute{Concordia University, Montreal QC, Canada \\
\and
Daesys Inc., Montreal QC, Canada \\
\and 
University of Waterloo, Waterloo ON, Canada\\
}
\maketitle              
\begin{abstract}
Shapley value attribution (SVA) is an increasingly popular explainable AI (XAI) method, which quantifies the contribution of each feature to the model’s output. However, recent work has shown that most existing methods to implement SVAs have some drawbacks, resulting in biased or unreliable explanations that fail to correctly capture the true intrinsic relationships between features and model outputs. Moreover, the mechanism and consequences of these drawbacks have not been discussed systematically. In this paper, we propose a novel error theoretical analysis framework, in which the explanation errors of SVAs are decomposed into two components: observation bias and structural bias. We further clarify the underlying causes of these two biases and demonstrate that there is a trade-off between them. Based on this error analysis framework, we develop two novel concepts: over-informative and under-informative explanations. We demonstrate how these concepts can be effectively used to understand potential errors of existing SVA methods. In particular, for the widely deployed assumption-based SVAs, we find that they can easily be under-informative due to the distribution drift caused by distributional assumptions. We propose a measurement tool to quantify such a distribution drift. Finally, our experiments illustrate how different existing SVA methods can be over- or under-informative. Our work sheds light on how errors incur in the estimation of SVAs and encourages new less error-prone methods.


\end{abstract}
\section{Introduction}

Explainable AI (XAI) is an emerging field that seeks to provide human-interpretable insights into complex and black-box machine learning (ML) models. Feature attribution, particularly SVA, is one increasingly popular XAI method, which explains a model's output by quantifying each input feature's contribution to the model \cite{lundberg2017unified,covert2020understanding}. The literature suggests that feature attribution methods can be \textit{true to the model} and/or \textit{true to the data} \cite{chen2020true,chen2023algorithms}. Feature attribution methods that are true to the model aim to understand the model's functional or algebraic dependencies on features. However, standard supervised ML learning models typically do not explicitly model dependencies between features \cite{watson2022rational,janzing2020feature}. Moreover, in the presence of feature interdependence, a model can often be written in different algebraic forms that perform identically \cite{frye2020shapley}. Hence, even if an attribution is exactly true to the model, it still might not correctly represent the intrinsic relationships between features and the model's output. If knowledge discovery is our objective, we want feature attributions to be \textit{true to the data}, representing the model's informational dependencies on features. Feature attribution methods that are true to the data put less emphasis on the particular model but more on the true underlying data-generating process \cite{chen2020true}. 
 
In this work, we focus on the study of SVAs that are true to the data. Since they can explain ML models more informatively, we call them \textit{informative SVAs}. In practice, SVAs have been widely used to assist decision explaining and model debugging. Moreover, researchers have recently begun applying SVAs for scientific discovery. For example, SVA techniques have been used to identify risk factors for diseases and mortality \cite{qiu2022interpretable,alatrany2024explainable,Cardiovascular2023,snider2021insights}; gain valuable new insights into genetic or molecular processes \cite{novakovsky2023obtaining,yagin2023explainable,janizek2021uncovering}; and capture informative patterns for fraud detection \cite{psychoula2021explainable}, etc. 

While SVAs provide promising directions to improve the understanding of underlying information systems, there remain concerns about their accuracy. Specifically, informative SVAs must be computed based on the true underlying distributions of the data, which are typically unknown in practice. Thus, we can only estimate these distributions using an observed dataset. However, the given dataset is usually too sparse to capture the complex distributions of high-dimensional or many-valued features, leading to significant estimation errors \cite{sundararajan2020many}. To address data sparsity, a number of approaches have been proposed \cite{aas2021explaining,Mase2019ExplainingBB,frye2020shapley,lundberg2018consistent}. Nevertheless, \cite{chen2023algorithms} and \cite{yeh2022threading} demonstrate that all of these approaches suffer from some drawbacks that lead to undesirable errors. Hence, in practice, instead of estimating the true distribution, most built-in SVA tools are designed based on some distributional assumptions, such as feature independence assumption. However, untenable assumptions may also result in incorrect attributions \cite{frye2020shapley}, making SVAs vulnerable to model perturbation \cite{slack2020fooling,lin2024robustness}. In this sense, most of the existing SVA methods are unreliable and error-prone. Furthermore, related works discussing errors of SVA methods are primarily method-specific and example-based \cite{aas2021explaining,frye2020shapley,Mase2019ExplainingBB,sundararajan2020many,yeh2022threading,slack2020fooling}. There has not been a comprehensive theoretical analysis of the errors of SVAs. Note that here we focus on the problems of SVA methods for discovering the informational dependencies between features and the target, while others consider causal relationships \cite{janzing2020feature,taufiq2023manifold} or the conceptual inadequacies of Shapley values for explanations \cite{kumar2020problems,kumar2021shapley,huang2023inadequacy}. 

In this paper, we establish a unified error analysis framework for SVAs. Under this framework, all explanation errors can be decomposed into two components: observation bias and structural bias. We analyze that observation bias arises due to the data sparsity, while structural bias results from unrealistic structural assumptions. We further demonstrate the trade-off between observation bias and structural bias. Based on this trade-off, we propose two novel concepts to describe SVAs: over-informativeness (with large observation bias) and under-informativeness (with large structural bias). Using our proposed error analysis framework, we theoretically analyze the potential over- and under-informativeness of various existing SVA methods. Furthermore, for the widely deployed distributional assumption-based SVA methods, we provide a mathematical analysis that shows how these methods can cause distribution drifts and produce under-informative explanations. To evaluate this risk, we propose a measurement tool to quantify the distribution drift.

We verify our theoretical error analyses on the Bike Sharing dataset \cite{misc_bike_sharing_dataset_275} and the Census Income dataset \cite{misc_adult_2}. The experimental results confirm our theoretical analysis that SVA methods that rely on structural assumptions tend to be under-informative, while excessive data smoothing methods can be sensitive to data sparsity, especially in low-density regions. This highlights the applicability of our error analysis framework, which can discern potential errors in many existing and future feature attribution methods.

\section{Background}

\subsection{Notation}

We seek to explain an ML model, denoted by $f: \mathcal{X} \rightarrow \mathcal{Y}$, which takes an instance \(x = (x_1, \ldots, x_d)\) from the domain set $\mathcal{X}$ as input and outputs predictions for a target variable \(Y \in \mathcal{Y} \subseteq \mathbb{R}\) (for classification, we typically focus on the predicted probability of a given class). In this paper, we use uppercase symbols \(X\), \(Y\) to denote random variables, and lowercase symbols \(x\), \(y\) to denote specific values. Furthermore, we use the notation \(X_S\) to refer to a sub-vector of \(X\) containing features in the subset \(S \subseteq [d] \equiv\{1, \ldots, d\}\), and \(X_{\bar{S}}\) to refer to its complementary sub-vector, which contains features from \(\bar{S} = [d] \setminus S\). We assume that \(X\) and \(Y\) follow an \textit{unknown} distribution \(p(X,Y)\). Instead of the true distribution, we are provided with a dataset $\mathcal{D}_p(X,Y) = \{(x^{(n)}, y^{(n)})\}_{n=1}^N$ of $N$ samples observed from $p(X,Y)$. This can be a training or testing set. Similarly, we use \(\mathcal{D}_p(X_S,Y)\) to denote the portion of $\mathcal{D}_p(X,Y)$ containing only features in the subset \(S\). Thus, sub-dataset \(\mathcal{D}_p(X_S,Y)\) is drawn from \(p(X_S,Y)\). 

A popular way to explain the model $f$ is to quantify each feature's contribution to a specific model output. This concept is referred to as \textit{feature attribution} and denoted by a vector \(\phi=(\phi_i, \ldots, \phi_d)\), where each $\phi_i$ is called the \textit{attribution score} or \textit{importance score} of feature $i$. The model output could be either an individual prediction \(f(x)\) for a specific sample \(x\), or a performance metric \(\mathbf{M}(f,\mathcal{D}_p(X,Y))\) evaluated across the entire dataset $\mathcal{D}_p(X,Y)$. In the former case, we term \(\phi\) as \textit{local feature attribution}, whereas in the latter case, \(\phi\) is referred to as \textit{global feature attribution}. 

\subsection{Informative SVA}
\label{S2.2}

Shapley value was originally a method from game theory to allocate credit to players in cooperative games \cite{shapley1953value}. They have been recently utilized to summarize each feature's contribution in model outputs \cite{lundberg2017unified,covert2020understanding}. Specifically, using Shapley values, each feature $i$'s importance score can be calculated as
\begin{equation}
\phi_i(v) = \sum_{S \subseteq [d] \setminus \{i\}} \pi(S) \left(v(S \cup \{i\}) - v(S) \right), \quad \textrm{ where } \; \pi(S) =  \frac{|S|!(d-|S|-1)!}{d!}, 
\label{equ:shap}
\end{equation}
and \(v(S): \mathcal{P}([d]) \rightarrow \mathbb{R}\) is a set function representing the model's output when only features in subset $S \subseteq [d]$ are considered. This importance score captures the average \textit{marginal contribution}, \(v(S \cup \{i\}) - v(S)\), of feature $i$ across all possible subsets of features $S$ that excludes $i$. 

SVA can be characterized under the framework of \textit{removal-based explanations} \cite{covert2021explaining}. Specifically, to design a SVA algorithm (also called a Shapley value \textit{explainer}), we need to specify two components: 
\begin{itemize}
    \item A \textbf{removal function (RF)} \(f_S (x_S)\) that can make predictions based on a sub-vector of input \(x_S\) instead of the full input vector \(x\).
    \vspace{2mm}
    \item A \textbf{value function} \(v_{f_S}(S)\) associated with the selected RF $f_S$. For example, for local feature attributions, we specify the value function as \(v_{f_S}(S)=f_S(x_S)\), while for global feature attributions, the value function can be designed as \(v_{f_S}(S)=\mathbf{M}(f_S, \mathcal{D}_p(X_S, Y))\) (see more discussions in \cite{covert2021explaining}).
\end{itemize}

\noindent
Under the removal-based framework, the RF \(f_S\) is leveraged to assess the impact of removing features in the complement subset \(\Bar{S}\) from the original model \(f\). Thus, the choice of RF significantly influences the resulting feature attributions. Recent research \cite{covert2021explaining,chen2020true,chen2023algorithms,covert2020understanding} emphasize that, to ensure the SVAs \textit{faithfully} capture the informational dependencies between model outputs and input features, we should select $f_S(x_S)$ to be the conditional expectation of model prediction $f(X)$ given the feature sub-vector $X_S = x_S$. Mathematically, 
\begin{align}
    f_S (x_S) & = \mathbb{E} [f(X)|X_S=x_S] = \mathbb{E}_{p(X_{\Bar{S}}|X_S=x_S)} [f(x_S, X_{\Bar{S}})] \label{eq:conditional-RF}.
\end{align}
In this case, we call $f_S$ the \textit{conditional RF}, and $\phi(v_{f_s})$ the \textit{informative SVA}. Since the true distribution $p(X)$ is typically unknown, the conditional distribution $p(X_{\Bar{S}}|X_S=x_S)$ is unavailable. Therefore, we can only \textit{estimate} $f_S(x_S)$ using the given dataset $\mathcal{D}_p(X)$ (which we call the \textit{explaining set}), such as the training set or testing set. There are two main challenges associated with this estimation task:
\begin{enumerate}
    \item \textbf{NP-hard} The exact computation of the SVA in Equation \eqref{equ:shap} requires the estimation of $f_S$ for all possible subsets $S$, which has exponential complexity in dimension $d$ \cite{aas2021explaining}. 
    \vspace{2mm}
    \item \textbf{Data sparsity} For each $f_S(x_S)$, we need to estimate the conditional distribution $p(X_{\Bar{S}}|X_S = x_S)$. However, in the explaining set, there could be very few or even no samples that match the condition $X_S = x_S$. This problem usually happens in problems that involve high-dimensional or many-valued features \cite{sundararajan2020many,chen2023algorithms}. For example, within a "bank dataset", it is unlikely to find any individual that exactly satisfies the condition: "$\mathtt{credit\_score}=3.879$, $\mathtt{income}=\$112,643$". 
\end{enumerate}

Various methods have been proposed to estimate the conditional RF $f_S$ (see discussion in \cite{covert2021explaining}), which either smooths the data or makes distributional assumptions \cite{sundararajan2020many}. However, due to the above two challenges, almost all existing estimation methods are error-prone and possibly computationally expensive, leading to incorrect explanations (see discussion in \cite{chen2023algorithms}). To gain better insights into this problem, in this paper, we provide a comprehensive analysis of potential explanation errors when estimating the informative SVA $\phi(v_{f_s})$.

\section{Observation Bias \& Structural Bias Trade-Off}


The SVA in Equation \eqref{equ:shap} is a function of the value function $v(S)$. Furthermore, the value function is intrinsically related to the RF. As a result, errors in estimating the conditional RF will directly cause errors in evaluating the value function, leading to errors in SVAs.

\subsection{Overfitting and Underfitting of the RF}

We use the notation $\hat{f}^{(N)}_S$ to denote an estimated conditional RF based on an explaining set of size $N$. Let \(\hat{f}_S = \lim_{N\to\infty} \hat{f}^{(N)}_S\) be the limit of the estimate when using an infinitely large explaining set. For instance, Frye et al. \cite{frye2020shapley} proposed adopting a supervised surrogate model $h_{\theta}(x_S)$ for the estimation of the conditional RF \(f_S(x_S)\). In this case, \(\hat{f}^{(N)}_S(x_S)=h_{\hat{\theta}^{(N)}}(x_S)\) and \(\hat{f}_S(x_S) = h_{\theta^*}(x_S)\), where $\hat{\theta}^{(N)}$, $\theta^*$ are obtained by minimizing the empirical MSE and true MSE, respectively. In essence, $\hat{f}^{(N)}_S$ is an estimate of $\hat{f}_S$, and $\hat{f}_S$ is a proxy for the true conditional RF $f_S$.

The error associated with an estimated RF $\hat{f}^{(N)}_S$ can be decomposed into two components: the \textit{estimation error} and the \textit{approximation error} \cite{shalev2014understanding}, expressed as:
\begin{equation}
\begin{split}
     \hat{f}^{(N)}_S - f_S & = (\hat{f}^{(N)}_S - \hat{f}_S) + (\hat{f}_S - f_S) \\
     & = \epsilon_{estimation} + \epsilon_{approximation}. \label{eq:decompose-f}
\end{split}
\end{equation}
The estimation error quantifies the risk of utilizing a finite dataset for the conditional RF estimation. This type of error can be highly sensitive to data sparsity but can be mitigated by either smoothing the data \cite{sundararajan2020many} or increasing the data size. The estimated RF $\hat{f}^{(N)}_S$ is said to be \textit{overfitting} at a point \(X_S=x_S\) if it exhibits a significant absolute estimation error \(|\hat{f}^{(N)}_S(x_S) - \hat{f}_S(x_S)|\).

On the other hand, the approximation error measures the level of risk associated with making distributional or modeling assumptions. In this case, the estimated RF $\hat{f}^{(N)}_S$ is said to be \textit{underfitting} at a point \(X_S=x_S\) if it demonstrates a significant absolute approximation error \(|\hat{f}_S(x_S) - f_S(x_S)|\). It is worth noting that underfitting cannot be alleviated through an increase in data size, but can be exacerbated by excessive data smoothing 

\subsection{Explanation Error Decomposition}

Since we use $\hat{f}^{(N)}_S$ to estimate the true conditional RF $f_S$, the true value function $v_{f_S}$ is estimated by $v_{\hat{f}_{S}^{(N)}}$. The difference between these two value functions causes explanation errors for the SVAs in Equation \eqref{equ:shap}. Using similar ideas as in Section 3.1, we propose to decompose the explanation error into 
\begin{equation}
\begin{split}
    \phi(v_{\hat{f}^{(N)}_S}) - \phi(v_{f_S}) & = \left(\phi(v_{\hat{f}^{(N)}_S}) - \phi(v_{\hat{f}_S})\right) + \left(\phi(v_{\hat{f}_S}) - \phi(v_{f_S})\right) \\
    & = \textit{observation bias} + \textit{structural bias}. \label{eq:decompose-phi}
\end{split}
\end{equation}

We call the first component \(\phi(v_{\hat{f}^{(N)}}) - \phi(v_{\hat{f}})\) the \textit{observation bias}, which occurs because we make explanations based on only a finite number of observations of the whole distribution. Next, we call the second component \(\phi(v_{\hat{f}}) - \phi(v_f)\) the \textit{structural bias}, arising from the utilization of an imperfect or limited knowledge structure to make explanations. While observation bias is caused by the estimation error, structural bias arises from the approximation error (see Equation \eqref{eq:decompose-f}).

\begin{figure}[tbh!]
\centering
\begin{tikzpicture}[
roundnode/.style={circle, draw=red!60, fill=red!5, thick, minimum size=11.5mm},
whitenode/.style={rectangle, draw=white!60, fill=white!5, thick, minimum height=2mm, minimum width=12mm},
ellipsenode/.style={ellipse, draw=red!60, fill=red!5, thick, minimum height=8mm, minimum width=16mm},
rectanglenode/.style={rectangle, draw=white!60, fill=red!10, thick, minimum height=6mm, minimum width=12mm},
]
\node[rectanglenode]   (1)    at (-4,0) {\small Observation Bias};
\node[rectanglenode]   (2)    at (-1,0) {\small Data Sparsity};
\node[rectanglenode]   (3)    at (2.5,0) {\small Structural Complexity};
\node[rectanglenode]   (4)    at (6.1,0) {\small Structural Bias};

\node[whitenode]   (1up)    at (-4,0.6) {\small $\Downarrow$};
\node[whitenode]   (2up)    at (-1,0.6) {\small $\Downarrow$};
\node[whitenode]   (3up)    at (2.5,0.6) {\small $\Downarrow$};
\node[whitenode]   (4up)    at (6.1,0.6) {\small $\Uparrow$};

\node[whitenode]   (1down)    at (-4,-0.6) {\small $\Uparrow$};
\node[whitenode]   (2down)    at (-1,-0.6) {\small $\Uparrow$};
\node[whitenode]   (3down)    at (2.5,-0.6) {\small $\Uparrow$};
\node[whitenode]   (4down)    at (6.1,-0.6) {\small $\Downarrow$};

\draw[dashed, -stealth] (1up) -- (2up) node[midway, above] {need to};
\draw[dashed, -stealth] (2up) -- (3up) node[midway, above] {need to};
\draw[ultra thin, -stealth] (3up) -- (4up) node[midway, above] {lead to};

\draw[dashed, -stealth] (4down) -- (3down) node[midway, below] {need to};
\draw[ultra thin, -stealth] (3down) -- (2down) node[midway, below] {lead to};
\draw[ultra thin, -stealth] (2down) -- (1down) node[midway, below] {lead to};

\end{tikzpicture}
\caption{An illustration of the trade-off between observation bias and structural bias. On the one hand, to reduce observation bias, it is necessary to alleviate the data sparsity, which requires us to decrease the structural complexity of the conditional RF approximation. However, this simplification of structural complexity might concurrently lead to an increase in structural bias. On the other hand, to reduce structural bias, we may need to increase the structural complexity, which inevitably entails an aggravation of the data sparsity, consequently increasing the observation bias.}
\label{fig:trade-off}
\end{figure}
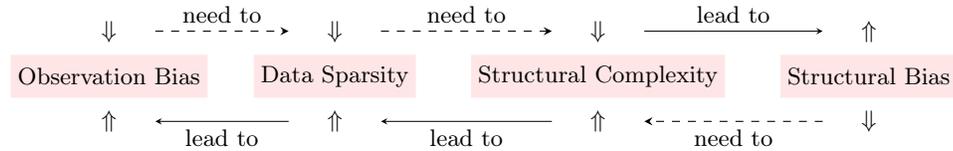

Observation bias may become substantial when the explaining set is too sparse to accurately capture the complex underlying distribution. To mitigate this, we can make simplifying structural assumptions to approximate $f_S$, for example, by using a surrogate model or an assumed distribution. However, imposing assumptions may cause the approximation to be inadequate. For example, using a surrogate model $h_{\theta}(x_S)$ with complexity $|\theta|$ may be insufficient to encompass a perfect $\theta^*$ that satisfy $h_{\theta^*}=f_S$. Moreover, making unrealistic distributional assumptions may drift the true underlying distribution $p(X)$ to a different one $q(X)$. Therefore, there is typically a trade-off between observation bias and structural bias in estimating the conditional RF using a finite explaining set. Figure \ref{fig:trade-off} gives an illustration of this trade-off. 

\subsection{Over-informative Explanation}

When the absolute value of observation bias \(|\phi(v_{\hat{f}^{(N)}_S}) - \phi(v_{\hat{f}_S})|\) is large, we say that the corresponding SVA is \textit{over-informative}. Over-informativeness often manifests in high-dimensional data and low-density regions, where the provided explaining set is typically too sparse to represent the whole population. Consequently, the estimated RF \(\hat{f}^{(N)}_S\) can easily be overfitting, resulting in an undesirable observation bias. When the SVA is over-informative, it may erroneously assign importance to uninformative or noisy features. To better illustrate the concept of over-informative SVAs, we present a toy example on two-dimensional data below.

\begin{example}[Over-informative SVA]
    Consider model \(f(x_1,x_2)=10x_2\) based on two independent features, \(X_1\) and \(X_2\). Suppose \(X_1 \sim \mathcal{N}(0,1)\) and \(X_2 \sim \mathcal{N}(0,1)\). Now, consider the case where we do not know the true distribution of \((X_1, X_2)\), and we only observe a dataset of 100 samples \(\{(x_1^{(1)}, x_2^{(1)}), \ldots, (x_1^{(100)}, x_2^{(100)})\}\). Suppose this dataset contains an outlier $(x_1, x_2) = (5,1)$, where the value $X_1=5$ is notably greater than that of all other samples. The objective is to explain the prediction \(f(5,1)=10\). According to the Shapley value formula in Equation \eqref{equ:shap}, in order to obtain feature attribution $\phi$, we need to estimate the conditional RFs $f_{\{\emptyset\}}, f_{\{1\}}, f_{\{2\}}$. Let us consider the empirical estimates \cite{sundararajan2020many} of these conditional RFs at $(5,1)$, which are:
    \begin{align*}
        \hat{f}^{(100)}_{\{\emptyset\}}(x_{\emptyset}) & = \frac{1}{100}\sum_{i=1}^{100} f(x_1^{(i)}, x_2^{(i)}) = \frac{1}{100}\sum_{i=1}^{100} 10x_2^{(i)} \approx 0, \\
        \hat{f}^{(100)}_{\{1\}}(5) & = \frac{\sum_{i=1}^{100}\mathbb{I}(x_1^{(i)} = 5)f(x_1^{(i)}, x_2^{(i)})}{\sum_{i=1}^{100}\mathbb{I}(x_1^{(i)} = 5)} = 10, \\ 
        \hat{f}^{(100)}_{\{2\}}(1) & = \frac{\sum_{i=1}^{100}\mathbb{I}(x_2^{(i)} = 1)f(x_1^{(i)}, x_2^{(i)})}{\sum_{i=1}^{100}\mathbb{I}(x_2^{(i)} = 1)} = 10.
    \end{align*}
    With these estimates, using Equation \eqref{equ:shap}, we can calculate \(\hat{\phi}_1 \approx 5\). This implies that $X_1$ contributes half to the prediction $f(5,1) = 10$. However, it is clear that, in reality, $X_1$ is an uninformative feature for $f$ and $\phi_1$ should always be 0. This error occurs because we observe only one sample with $X_1 = 5$ in the dataset, making the empirical estimator $\hat{f}^{(100)}_{\{1\}}$ overfitting at $(5,1)$. Since the true conditional RF is $f_{\{1\}} = 0$, the estimation error is 10, causing the observation bias to be 5. In this case, the SVA score $\hat{\phi}_1$ is over-informative and it erroneously assigns importance to irrelevant features.
\end{example}

\subsection{Under-informative Explanation}
Conversely, when the absolute value of structural bias \(|\phi(v_{\hat{f}_S}) - \phi(v_{f_S})\) is large, we say that the corresponding SVA is \textit{under-informative}. In practice, making unreasonable assumptions is the primary reason for under-informativeness. When the SVA is under-informative, it may underestimate or even ignore some relevant mutual information between input features and model outputs. For example, Chen et al. \cite{chen2020true} demonstrate that assuming feature independence can result in highly correlated features receiving considerably different importance scores. We give a toy two-dimensional example below to illustrate an under-informative SVA.

\begin{example}[Under-informative SVA]
    Suppose we are given two features $X_1$ and $X_2$, where \(X_1=2 X_2\), representing the same factor in two different units, e.g., price in different currencies or temperature in different scales. Consider two linear models \(f_1(x_1, x_2)=10x_1+x_2\) and \(f_2(x_1, x_2)=x_1+19x_2\), which both equals $21x_2$. In essence, $f_1$ and $f_2$ are the same model with different algebraic forms. However, under the feature independence assumption, they can be explained in two different ways. Assume $\mathbb{E}[X_1] = \mathbb{E}[X_2] = 0$ and suppose we are interested in explaining the same prediction \(f_1(2,1)=f_2(2,1)=21\). Using the SVA formula for linear models under independent feature assumptions\footnote{Following \cite{lundberg2017unified}, given a linear model $f(x) = \sum_{j=1}^d\beta_jx_j + \beta_0$, under the feature independence assumption, the SVA for the $j$th feature can be calculated as $\phi_j = \beta_j(x_j - \mathbb{E}[X_j])$.}, we can calculate \(\hat{\phi}_1=20, \hat{\phi}_2=1\) for $f_1$, and \(\hat{\phi}_1=2, \hat{\phi}_2=19\) for $f_2$. That means $X_1$ is given dominantly high feature attribution for $f_1$ while $X_2$ is given dominantly high feature attribution for $f_2$. In reality, $X_1$ and $X_2$ should receive the same attribution score, i.e., \(\phi_1=\phi_2\), because they provide the same information. In this case, both explanations are under-informative due to the unrealistic feature independence assumption.
\end{example}

In summary, SVA could be over-informative if it is estimated based on insufficient observations. Meanwhile, it could also be under-informative if it is approximated based on unrealistic structural assumptions. In the following sections, we use the error analysis framework proposed in Equation \eqref{eq:decompose-phi} to analyze the over- and under-informativeness of existing conditional RF estimation methods. These methods can be categorized into two main approaches: smoothing the data and making distributional assumptions. 

\subsection{Explanation Error Analysis of Data-Smoothing Methods} \label{sec:error-data-smoothing}

To address the challenge of data sparsity, one effective method is to smooth the explaining set. Typically, the data can be smoothed using either non-parametric kernel-based approaches or parametric model-based approaches. However, excessive data smoothing can lead to serious structural bias. Unfortunately, it is unclear to what extent the explaining set should be smoothed \cite{sundararajan2020many}. Below we analyze the potential explanation errors of some popular data smoothing methods. 

\paragraph{Empirical conditional RF}\cite{sundararajan2020many}: the structural bias is zero because the empirical estimator will converge to the true conditional RF when the data size goes to infinity. However, the empirical conditional RF is usually seriously over-informative when data sparsity exists (as illustrated in Example 1). 

\paragraph{Non-parametric kernel-based approaches}\cite{aas2021explaining,Mase2019ExplainingBB}: in this type of approach, the extent of data smoothing is controlled by the bandwidth(s) of the kernel, which could be set either too conservatively, resulting in over-informativeness, or too generously, leading to under-informativeness. Moreover, the selected kernel function might not correctly define the similarity between samples \cite{chen2023algorithms}, causing undesirable structural bias.

\paragraph{Parametric model-based approaches} \cite{frye2020shapley}: for both the conditional generative model and supervised surrogate model proposed in \cite{frye2020shapley}, the extent of data smoothing is controlled by the complexity of the selected neural networks. Over-informativeness and under-informativeness respectively coincide with the overfitting and underfitting of the trained neural network. However, controlling the overfitting and underfitting of this trained neural network is challenging. First, since the neural network is trained on an exponential number of all possible sub-datasets \(\mathcal{D}_p(X_S)\), it is sometimes difficult to ensure learning optimality within an acceptable computation time \cite{chen2023algorithms}. As a result, non-optimal learning may result in structural bias. Furthermore, even if a neural network is well-trained, it might still be overfitting under data sparsity in low-density regions (see examples in \cite{yeh2022threading}), causing observational bias. 

\paragraph{TreeSHAP}\cite{lundberg2020local2global,lundberg2018consistent}: this is a specific SVA method for tree-structured models. TreeSHAP is usually under-informative. First, it utilizes the predefined tree structure of the original model, which was trained under unclear assumptions about feature dependencies \cite{aas2021explaining}. Second, it approximates the conditional expectation \(\mathbb{E} [f(X)|X_S=x_S]\) by averaging the predictions from all leaves that are not against the condition \(X_S=x_S\). Essentially, this procedure relaxes the condition \(X_S=x_S\) into a set of weaker conditions. For instance, with a stump containing two leaves "\(X_1 < 10\)" and "\(X_1 \ge 10\)", we approximate \(\mathbb{E} [f(X)|X_1=8]\) by \(\mathbb{E} [f(X)|X_1<10]\). This relaxation of conditions introduces structural bias.

\subsection{Explanation Error Analysis of Distributional Assumptions-Based Methods} \label{sec:error-assumption-based}

Besides smoothing the data, an alternative way to mitigate data sparsity is to approximate the conditional distribution \(p(X_{\Bar{S}}|X_S=x_S)\) with an assumed distribution $r(X_{\Bar{S}})$. In this paper, we call  $r(X_{\Bar{S}})$ the \textit{removal distribution}, as it is the assumed distribution for removed feature subset $X_{\Bar{S}}$. As discussed in \cite{chen2023algorithms}, there are four common removal distributions: 
\begin{enumerate}
        \item \textit{Baseline}: $r(X_{\Bar{S}})=\mathbbm{1}(X_{\Bar{S}}=x^b_{\Bar{S}})$, assuming \(X_{\bar{S}}\) has a constant value $x^b_{\Bar{S}}$ \cite{sundararajan2020many}.
        \vspace{2mm}
        \item \textit{Marginal}: $r(X_{\Bar{S}})=p(X_{\Bar{S}})$, assuming \(X_S\) and \(X_{\bar{S}}\) are independent \cite{lundberg2017unified}.
        \vspace{2mm}
        \item \textit{Product of marginal}: $r(X_{\Bar{S}})=\prod_{i \in \bar{S}} p(X_i)$, assuming each feature in $\Bar{S}$ is independent \cite{datta2016algorithmic}. 
        \vspace{2mm}
        \item \textit{Uniform}: $r(X_{\Bar{S}})=\prod_{i \in \bar{S}} u_i(X_i)$, where $u_i$ denotes a uniform distribution over $\mathcal{X}_i$. In this case, each feature in $\Bar{S}$ is assumed to be independently and uniformly distributed \cite{strumbelj2010efficient}.
\end{enumerate}

With \(p(X_{\Bar{S}}|X_S=x_S) \approx r(X_{\Bar{S}})\), the conditional RF $f_S$ in formula \eqref{eq:conditional-RF} can be approximated as 
\begin{equation}
    \hat{f}_S(x_S) = \mathbb{E}_{r(X_{\Bar{S}})} [f(x_S, X_{\Bar{S}})] = \int f(x_S, x'_{\Bar{S}}) r(X_{\Bar{S}}=x'_{\Bar{S}}) d x'_{\Bar{S}},
\label{eq:assumption-based-empirical-RF}
\end{equation}
which can be empirically estimated by 
\begin{equation}
    \hat{f}^{(N)}_S(x_S) = \frac{1}{N} \sum_{n=1}^N f(x_S, x_{\bar{S}}^{(n)}), \label{eq:assumption-based-empirical-RF-estimator}
\end{equation}
using an explaining set $\mathcal{D}_r(X)=\{(x^{(n)})\}_{n=1}^N$ drawn from $r(X)$. 

\paragraph{Observational bias:} The purpose of making assumptions is to reduce the distribution complexity, and thus the observation bias. In particular, to estimate the conditional distribution $p(X_{\Bar{S}}|X_S = x_S)$ for any arbitrary $x_S$, we require a dataset with complexity $O( |\mathcal{X}|)$. This complexity will change when using an assumed removal distribution $r(X_{\Bar{S}})$. Table \ref{tab:complexity-r-func} summarizes the data complexity requirement for the above four removal distributions. 

\begin{table}[tbh!]
\caption{The complexity of different removal functions.} 
\begin{center}
\begin{tabular}{|c|c|c|}
\hline
\textbf{Removal distribution} & \hspace{10mm} \textbf{Formula} \hspace{10mm} & \textbf{Data complexity required} \\
\hline
Conditional & $p(X_{\Bar{S}}|X_S = x_S)$ & $O(|\mathcal{X}|)$ \\
Baseline & $\mathbbm{1}(X_{\Bar{S}}=x^b_{\Bar{S}})$ & $O(1)$ \\
Marginal & $p(X_{\Bar{S}})$ & $O(|\mathcal{X}_{\Bar{S}}|)$ \\
Product of marginals & $\prod_{i\in \Bar{S}}p(X_i)$ & $O\left(\prod_{i \in \bar{S}} |\mathcal{X}_i|\right)$ \\
Uniform & $\prod_{i \in \bar{S}} u_i(X_i)$ & $O\left(\prod_{i \in \bar{S}} |\mathcal{X}_i|\right)$ \\
\hline
\end{tabular}
\end{center}
\label{tab:complexity-r-func}
\end{table}

\noindent
From Table \ref{tab:complexity-r-func}, we can see that the baseline removal distribution simplifies the conditional distribution into a constant value, thus having a zero observation bias. The marginal removal distribution also decreases the data complexity requirement from $O(|\mathcal{X}|)$ into $O(|\mathcal{X}_{\Bar{S}}|)$. 
However, not all the distributional assumptions can ensure a decrease in complexity, even though the assumptions are strong. For example, both product of marginal and uniform removal distributions require a dataset with a complexity of $O\left(\prod_{i \in \bar{S}} |\mathcal{X}_i|\right)$, which might not be necessarily lower than the complexity requirement of conditional distribution (i.e., $O(|\mathcal{X}|)$) in the presence of dependencies among features.

\paragraph{Structural bias:} By reducing the data complexity requirement, making some distributional assumptions can reduce the observation bias. However, if these assumptions are far from the true underlying distribution, they could also engender considerable structural bias. Specifically, distributional assumptions can make the true joint distribution \(p(X)\) drift towards a different distribution \(q(X)\), where $q(X_{\bar{S}}|X_S=x_S)=r(X_{\bar{S}})$. To analyze the structural bias induced by distributional drift, we introduce the following definitions.

\begin{definition}
    A sample $x$ is defined as an \textbf{out-of-distribution (OOD) sample} of \(p(X)\), denoted as $x \notin p(X)$, if $p(X = x) = 0$. Conversely, if $p(X = x) > 0$, it is defined as an \textbf{in-distribution sample} of \(p(X)\), denoted as $x \in p(X)$.
\end{definition}
\begin{definition}
    The \textbf{OOD rate} of \(q(X)\) to \(p(X)\) is defined as the proportion of samples drawn from $q(X)$ that are OOD samples of \(p(X)\), denoted as \(\mathbf{Pr}\{X \notin p(X)|X \in q(X)\}\).
\end{definition}

\noindent
For an arbitrary value \(x_S\) observed from $p(X_S)$, the instance $x = (x_S, x'_{\Bar{S}})$ where $x'_{\Bar{S}} \sim r(X_{\Bar{S}})$ is called a hybrid sample \cite{chen2023algorithms}. As a result of the distribution drift, hybrid samples \((x_S, x'_S) \sim q(X)\) could be either in-distribution or OOD samples of \(p(X)\). Thus, we can derive the approximation error of the conditional RF estimator $\hat{f}_S(x_S)$ in Equation \eqref{eq:assumption-based-empirical-RF} as
\begin{align*}
    & \hat{f}_S(x_S) - f_S(x_S) \\
  = & \int_{(x_S,x'_{\bar{S}}) \in q(X)} f(x_S,x'_{\bar{S}}) r(X_{\bar{S}}=x'_{\bar{S}}) dx'_{\bar{S}} - f_S(x_S) \\
  = & \int_{(x_S,x'_{\bar{S}}) \notin p(X)} f(x_S,x'_{\bar{S}}) r(X_{\bar{S}}=x'_{\bar{S}}) dx'_{\bar{S}} \quad + \\
    &  \int_{(x_S,x'_{\bar{S}}) \in p(X)} f(x_S,x'_{\bar{S}}) r(X_{\bar{S}}=x'_{\bar{S}}) dx'_{\bar{S}} -f_S(x_S) \\
  = &  \int_{(x_S,x'_{\bar{S}}) \notin p(X)} f(x_S,x'_{\bar{S}}) r(X_{\bar{S}}=x'_{\bar{S}}) dx'_{\bar{S}} \quad +  \\
    & \int_{(x_S,x'_{\bar{S}}) \in p(X)} f(x_S,x'_{\bar{S}}) \left[r(X_{\bar{S}}=x'_{\bar{S}}) - p(X_{\bar{S}}=x'_{\bar{S}}|X_S=x_S)\right] dx'_{\bar{S}}. \label{eq:approx-err-decomposition} \numberthis
\end{align*}
Therefore, the approximation error of assumption-based RFs stems from two sources: (i) the inclusion of OOD samples in the approximation; and (ii) changes in the probability density of in-distribution samples. The OOD sample-related approximation error may contribute to a large proportion of structural bias, especially when the OOD rate is high. In practice, some OOD samples may be senseless. For instance, the OOD samples could represent a bank client who is 20 years old but has 25-year working experience, or a clinic patient whose systolic blood pressure is lower than his diastolic blood pressure. Moreover, adversarial attacks have been designed in the literature \cite{slack2020fooling} to arbitrarily manipulate model explanations (feature attributions). Under our error analysis framework, it is easy to see that these attacks essentially target the OOD sample-related approximation error in Equation \eqref{eq:approx-err-decomposition}, intentionally modifying the structural bias. 

\section{OOD Measurement of Distribution Drift}

In practice, assumption-based RFs, such as the baseline RF and marginal RF, are widely used thanks to their simple implementations \cite{lin2024robustness}. For these methods, explanation errors mainly arise from structural bias caused by distributional assumptions, which are unchangeable once the assumptions are made. Hence, evaluating structural bias or under-informativeness resulting from distributional assumptions is crucial. However, it is impossible to directly measure the structure bias because the true conditional RF $f_S$ is unknown. As discussed in Section \ref{sec:error-assumption-based}, structural bias arises from distribution drift, which usually leads to the use of OOD samples in estimating SVAs. Therefore, we can alternatively assess structural bias or under-informativeness by measuring how much the distribution drifts, and how high the OOD rate is. 

\subsection{Distribution Drift \& OOD Detection}
\label{sec:OOD-classifier}

Let $\mathbf{S}$ be a random variable on domain $\mathcal{P}([d]) \setminus [d]$ (i.e., the power set of $[d]$ excluding $[d]$, which is the set of all possible subsets involved in the computation of SVA scores for all $d$ features). 

\begin{lemma}
    For each $S \in \mathcal{P}([d]) \setminus [d]$, \(\mathbf{Pr}\{\mathbf{S}=S \} = \frac{1}{d \cdot  \binom{d}{|S|}} \).
\end{lemma}

\begin{proof}
    According to Equation \eqref{equ:shap}, the SVA score $\phi_i$ is essentially the weighted average of feature $i$'s marginal contribution over all possible subsets $S \subseteq [d] \setminus \{i\}$, with weights equal $\pi(S)$. In the context of all $d$ features, a subset $S$ only appears when computing SVA scores for features that are not in $S$. There are $d-|S|$ such features. Therefore, the probability function of $\mathbf{S}$ can be derived as
    \begin{equation}
        \mathbf{Pr}\{\mathbf{S}=S \} = \frac{d-|S|}{d} \pi(S) = \frac{d-|S|}{d} \cdot \frac{|S|!(d-|S|-1)!}{d!} = \frac{1}{d \cdot  \binom{d}{|S|}}. \nonumber \qquad \square
    \end{equation}
\end{proof}

Given $\mathbf{S}=S$ and an instance $x$, we have
\[p(X = x|\mathbf{S}=S) = p(X_S = x_S)p(X_{\Bar{S}} = x_{\Bar{S}}|X_S = x_S).\]
By assuming a removal distribution \(r(X_{\bar{S}})\) on the conditional distribution $p(X_{\Bar{S}} = x_{\Bar{S}}|X_S = x_S)$, the distribution drift into 
\begin{equation}
    q\left(X=x|\mathbf{S}=S \right) = p(X_S=x_S) r(X_{\bar{S}}=x_{\bar{S}}). 
\end{equation}
Then, considering all possible subsets $S$, the marginal density of a hybrid sample $x \sim q(X)$ can be computed as 
\begin{equation}
    q\left(X=x \right) = \frac{1}{d} \sum_{S \in \mathcal{P}([d]) \setminus [d]} \frac{1}{\binom{d}{|S|}} p(X_S=x_S) r(X_{\bar{S}}=x_{\bar{S}}).
\end{equation}
If the assumed removal distribution \(r(X_{\bar{S}}) \neq p(X_{\bar{S}}|X_S=x_S)\), there will be a distribution drift from $p(X)$ to $q(X)$. For example, when using baseline and marginal removal distributions, the true distribution $p(X)$ could drift into $q^{baseline}(X)$ and $q^{marginal}(X)$, respectively, where
\begin{align}
 q^{baseline}(X) & = \frac{1}{d} \sum_{S \in \mathcal{P}([d]) \setminus [d]} \frac{1}{\binom{d}{|S|}} p(X_S) \mathbbm{1}(X_{\Bar{S}}=x^b_{\Bar{S}}), \qquad \text{and} \label{q:baseline} \\
 q^{marginal}(X) & = \frac{1}{d} \sum_{S \in \mathcal{P}([d]) \setminus [d]} \frac{1}{\binom{d}{|S|}} p(X_S) p(X_{\bar{S}}). \label{q:marginal}
\end{align}

To detect the OOD samples, Slack et al. \cite{slack2020fooling} proposed training a binary classifier $\mathtt{ood\_score}(x)$ to predict whether a given sample \(x\) belongs to \(p(X)\) or \(q(X)\). Specifically, we first generate a $M$-size dataset \(\mathcal{D}_q(X)\) from \(q(X)\) and label it as 0. This dataset is then combined with the provided explaining set \(\mathcal{D}_p(X)\) labeled as 1 to train the classifier. The classifier returns an OOD score, approximating the probability that the input $x$ comes from $p(X)$. A hybrid sample $(x_S, x'_S)$ is considered an OOD sample if $\mathtt{ood\_score}(x_S, x'_S)$ is smaller than a selected threshold $t$. 

Furthermore, let \(C=\mathtt{ood\_score}(X)\) denote the OOD score random variable, and let \(p(C)\), \(q(C)\) denote the distributions of $C$ induced by \(p(X)\), \(q(X)\) respectively. If no distribution drift occurs, i.e., \(q(X)=p(X)\), then we have \(q(C)=p(C)\). Conversely, if \(q(C) \neq p(C)\), then \(q(X) \neq p(X)\), indicating a distribution drift. Thus, to detect the distribution drift, we propose comparing the distribution drift by examining the distributions of OOD scores $C$ calculated on \(\mathcal{D}_p(X)\) and \(\mathcal{D}_q(X)\). One possible way to compare the two distributions is to visualize their density histograms in a single plot (see Figure \ref{Fig:ood drift} for an example). Another way is to quantify the distribution drift by calculating the \textit{total variation distance} \cite{devroye2013probabilistic}:
\begin{equation}
    D_{TV} [p(C), q(C)] = \frac{1}{2} \int_0^1 | p(C=c) - q(C=c)| d c.
\end{equation}
The total variation distance can be conveniently estimated by half the absolute sum of density difference in all bins between the two density histograms.

\section{Experiments}
\label{S:5}

In this section, we conduct experiments to verify the error analyses we performed on existing SVA methods in previous sections. First, we demonstrate how to apply the method we proposed in Section \ref{sec:OOD-classifier} to detect and measure the distribution drifts caused by different distributional assumptions that have been used in the literature. Next, we will show that this distribution drift can lead to under-informative attributions, which assign significantly different important scores to highly correlated features. Finally, we demonstrate how data sparsity can cause over-informative attributions, which assign high important scores to irrelevant or noisy features.

\paragraph{Dataset} To assure the generalizability of our conclusions, we conduct our experiments on two datasets. Our first dataset is the Bike Sharing Dataset, which contains 17,389 records of hourly counts of bike rentals in 2011-2012 in the Capital Bike Sharing system \cite{misc_bike_sharing_dataset_275}. The dataset comprises a set of 11 features, following an unknown joint distribution. The objective is to predict the number of bikes rented during a specific hour of the day, based on various features related to time and weather conditions, such as hour, month, humidity, and temperature. The second dataset that we use is the Census Income (also known as Adult) dataset, which contains information such as age, work class, education, etc. of 48,842 adults \cite{misc_adult_2}. The goal is to predict whether an adult's income exceeds 50,000 dollars. The dataset is extracted from the 1994 Census database. In each dataset, samples with missing data are removed. 

For the Bike Sharing dataset, we aim to explain an xgBoost regressor trained on a training set of 15,379 samples and tested on a testing set of 2,000 samples. In addition, we split the Census Income dataset into a training set of 32,561 samples and a testing set of 4,000 samples. Our goal for the Census Income dataset is to explain an xgBoost classifier trained and tested on the respective sets.

\subsection{Distribution Drift Detection} \label{sec:exp-1}

In this section, we will demonstrate how different distributional assumptions caused distribution drifts and estimate the corresponding OOD rates. Besides the training and test datasets described above, we generate four sets of hybrid samples by using four different removal distributions: uniform, product of marginal, marginal, and baseline. To make the results comparable, we calculate the OOD scores of the four hybrid sample sets using a \textit{single} OOD classifier. Such an OOD classifier is trained using samples from the training set (labeled as 1) and hybrid samples generated from uniform removal distribution (labeled as 0). Note that this OOD classifier is still valid for OOD detection on hybrid samples generated from the other distributions because those samples are in-distribution of the uniform removal distribution. 

\begin{figure}[tbh!]
\centering
    \begin{subfigure}[b]{1\textwidth}
         \centering
         \caption{Bike Sharing Dataset}
         \includegraphics[width=\textwidth]{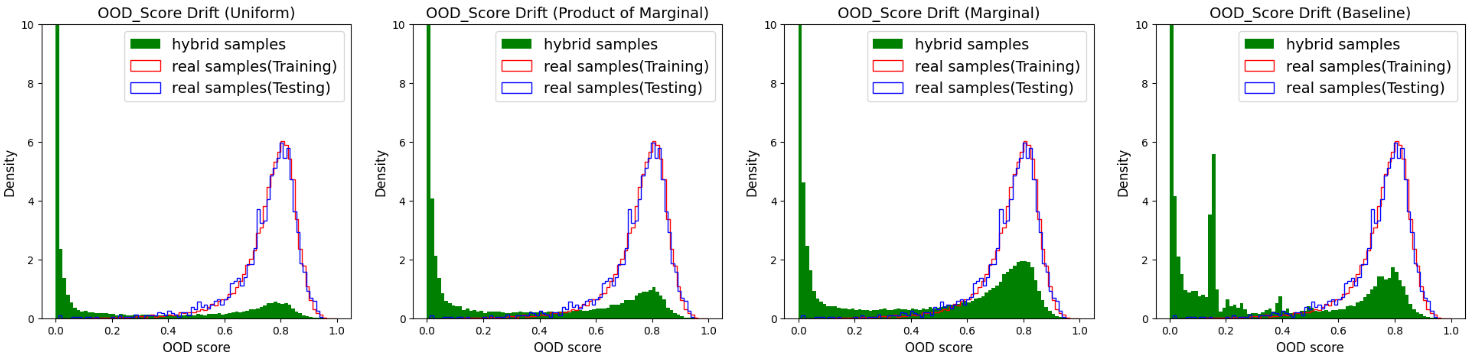}
         \label{fig:drift1}
    \end{subfigure}
    \hfill
    \begin{subfigure}[b]{1\textwidth}
         \centering
         \caption{Census Income Dataset}
         \includegraphics[width=\textwidth]{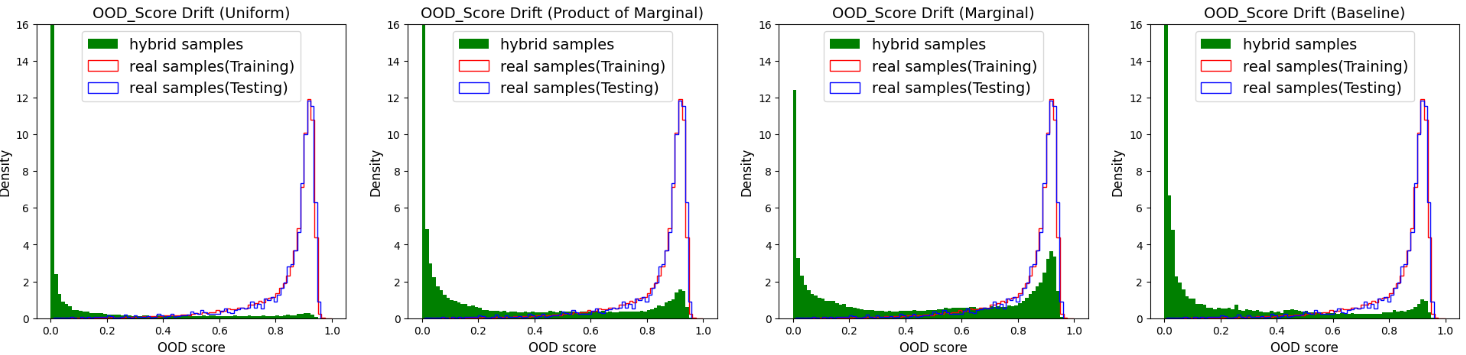}
         \label{fig:drift2}
    \end{subfigure}
    
\caption{The density histograms of OOD scores on real samples and hybrid samples}
\label{Fig:ood drift}
\end{figure}

\begin{table}[tbh!]
\caption{The OOD rates and total variance distance} 
\begin{center}
\begin{tabular}{|c|c|c|}
\hline
     \textbf{Removal distribution} & \textbf{OOD rate (t=0.3)} & \textbf{Total Variance Distance} \\
\hline
    \multicolumn{3}{|c|}{Bike Sharing Dataset}\\
\hline
    Uniform & 0.866 & 0.868  \\
    Product of Marginal  & 0.757 & 0.77 \\
    Marginal  & 0.538 & 0.578  \\
    Baseline  & 0.666 & 0.696 \\
\hline
    \multicolumn{3}{|c|}{Census Income Dataset}\\
\hline
    Uniform & 0.901 & 0.903  \\
    Product of Marginal  & 0.69 & 0.729 \\
    Marginal  & 0.448 & 0.524  \\
    Baseline  & 0.756 & 0.804 \\
\hline
\end{tabular}
\end{center}
\label{ood drift}
\end{table}

The trained OOD classifier is then used to calculate OOD scores $C$ for all real samples from both the training and testing sets, as well as for all hybrid samples in the four generated sets. We plot density histograms of these OOD scores in Figure~\ref{Fig:ood drift}. The total variance distances between the OOD score distributions calculated from the training samples versus the generated hybrid samples are given in Table~\ref{ood drift}). First, we observe that the OOD density histograms of the training and test samples overlap, which implies that there is no distribution drift detected between the training and testing sets of both datasets. Second, we observe that all four removal distributions introduce noticeable distribution drifts, together with a considerable number of OOD samples. This is particularly evident for the uniform and product of marginal removal distributions, where the OOD rates are exceptionally high when adopting a threshold of 0.3 (0.866 and 0.757 for the Bike Sharing dataset, and 0.901 and 0.69 for the Census Income dataset, respectively). In contrast, the marginal removal distribution seems to exhibit the least distribution drift, ($D_{TV} = 0.578$ in the Bike Sharing dataset and $D_{TV}=0.524$ in the Census Income dataset, respectively). Finally, the fact that the total variance distances are all greater than 50\% for all removal distributions in both datasets highlights the severity of the distribution drifts. 

\subsection{Under-informativeness Audit}

In Section \ref{sec:exp-1}, we showed that assumption-based methods caused severe distribution drifts. In this section, we will demonstrate that these distribution drifts can contribute to under-informative attributions. 

For both datasets, we explain model predictions on 100 samples using SVAs calculated from five different RFs, namely SHAP-B (with baseline RF), SHAP-M (with marginal RF), SHAP-PoM (with product of marginal RF), SHAP-U (with uniform RF) and SHAP-S (with surrogate model-estimated conditional RF). In addition, TreeSHAP is also used to explain the predictions of xgBoost models on each dataset.

\begin{figure}[tbh!]
\centering
    \begin{subfigure}[b]{0.48\textwidth}
         \centering
         \includegraphics[width=\textwidth]{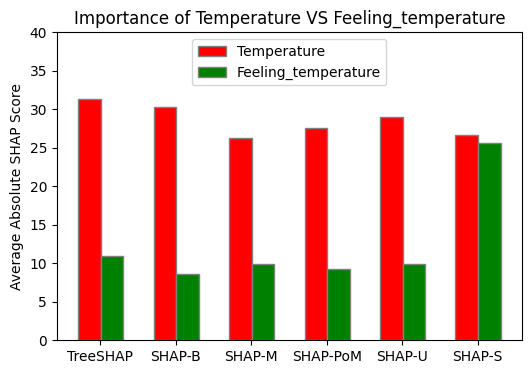}
         \caption{Bike Sharing Dataset}
         \label{fig:exp2-bike}
    \end{subfigure}
    \vspace{1mm}
    \begin{subfigure}[b]{0.48\textwidth}
         \centering
         \includegraphics[width=\textwidth]{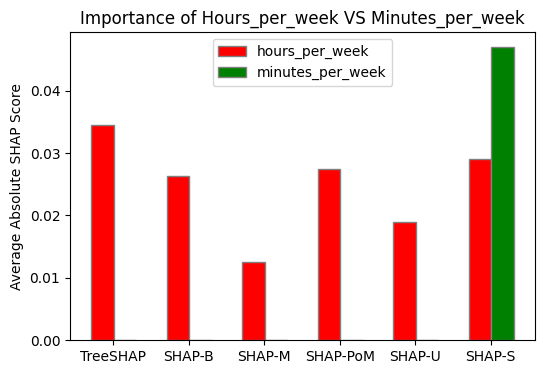}
         \caption{Census Income Dataset}
         \label{fig:exp2-income}
    \end{subfigure}
    
\caption{Under-informativeness audit on 100 predictions. (a) the average absolute SHAP scores of features "$\mathtt{Temperature}$" and "$\mathtt{Feeling\_Temperature}$" (ideally, they should receive similar scores); (b) the average absolute SHAP scores of features "$\mathtt{Hours\_per\_week}$" and "$\mathtt{Minutes\_per\_week}$" (ideally, they should receive the identical scores).}
\label{over and under test}
\end{figure}

Intuitively, an informative SVA method should (1) assign similar attribution scores to the two highly correlated features $\mathtt{Temperature}$ and $\mathtt{Feeling\_Temperature}$ with Pearson correlation of 0.99 for the Bike Sharing dataset as they convey almost the same information; (2) assign exactly the same attribution score to features "$\mathtt{Hours\_per\_week}$" and "$\mathtt{Minutes\_per\_week}$" for the Census Income Dataset because they hold the same information but in different scales.

From Figure~\ref{fig:exp2-bike}, we can observe that TreeSHAP, SHAP-B, SHAP-M, SHAP-PoM, and SHAP-U all assign much higher importance scores to feature $\mathtt{Temperature}$ than $\mathtt{Feeling\_Temperature}$. Moreover, in Figure~\ref{fig:exp2-income}, TreeSHAP, SHAP-B, SHAP-M, SHAP-PoM, and SHAP-U only assign importance to feature "$\mathtt{Hours\_per\_week}$" and ignore feature "$\mathtt{Minutes\_per\_week}$". This is because these methods do not consider the dependencies among features, leading to under-informative attributions. In contrast, SHAP-S trains a surrogate model to learn feature correlations, thus able to allocate similar importance scores to $\mathtt{Temperature}$ and $\mathtt{Feeling\_Temperature}$. For the Census Income dataset, even though SHAP-S mitigates the problem of under-informativeness by assigning importance to both "$\mathtt{Hours\_per\_week}$" and  "$\mathtt{Minutes\_per\_week}$", however, these scores are not the same. This indicates that the SHAP-S still produces structural bias and does not completely resolve the under-informativeness problem for the Census Income dataset.

\subsection{Over-informativeness Audit}

\begin{figure}[tbh!]
\centering
    \begin{subfigure}[b]{0.48\textwidth}
         \centering
         \includegraphics[width=\textwidth]{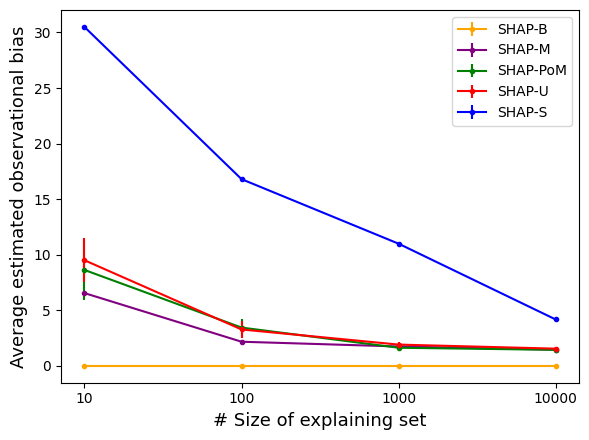}
         \caption{Bike Sharing Dataset}
         \label{fig:noisy feature}
    \end{subfigure}
    \vspace{1mm}
    \begin{subfigure}[b]{0.48\textwidth}
         \centering
         \includegraphics[width=\textwidth]{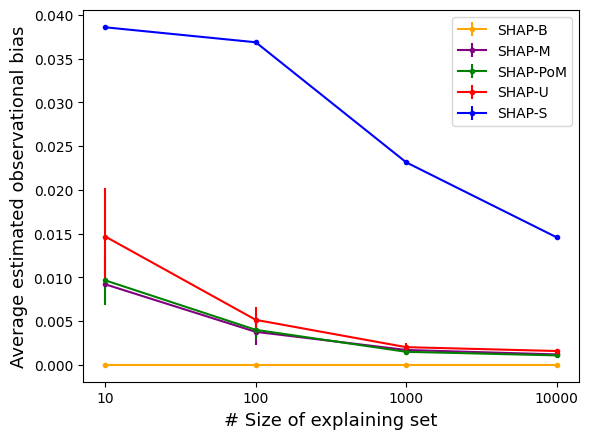}
         \caption{Census Income Dataset}
         \label{fig:temp}
    \end{subfigure}
    
\caption{The change in average estimated observation bias of the SVAs as the size of the explaining set changes.}
\label{fig:exp3}
\end{figure}

In this section, we turn our attention to over-informativeness and observation bias. Recall that, the observation bias in Equation \eqref{eq:decompose-phi} is $\phi(v_{\hat{f}_S^{(N)}}) - \phi(v_{\hat{f}_S})$ where $\hat{f}_S = \lim_{N\to\infty}\hat{f}_S^{(N)}$. However, since we do not have an infinite explaining set, we cannot evaluate the observational bias directly. In this experiment, we estimate $\hat{f}_S$ by $\hat{f}_S^{(M)}$, where $\hat{f}_S^{(M)}$ is estimated using the whole training sets of both datasets. That is, $M = 15,379$ for the Bike Sharing dataset and $M = 32,561$ for the Census Income dataset. For random explaining sets with $N \in \{10,100,1000,10000\}$, we estimate the average absolute observation bias in the SVAs of 100 predictions, namely

\begin{equation*}
    \frac{1}{100} \frac{1}{d} \sum_{i=1}^{10} \sum_{j=1}^{d} |\phi_{ij}(v_{\hat{f}^{(N)}_S}) - \phi_{ij}(v_{\hat{f}^{(M)}_S})|,
\end{equation*}

\noindent
where $\phi_{ij}$ is the SVA of the $j$th feature in the $i$th prediction. The results are plotted in Figure~\ref{fig:exp3}. We observe similar trends in both datasets. Generally, observation bias decreases when the size of the explaining set increases. This illustrates the relationship between observation bias and data sparsity. However, different methods exhibit different sensitivity to data sparsity. Specifically, SHAP-B always has 0 observation bias, which agrees with our analysis in Section \ref{sec:error-assumption-based}. For SHAP-M, SHAP-PoM, and SHAP-U, observation bias quickly stabilizes at $N = 1,000$. In contrast, SHAP-S shows high sensitivity to data sparsity, especially for the Census Income Data, at $N = 10,000$, the observation bias of SHAP-S is still much higher than those of other methods. Note that both datasets that we use contain less than 20 features. If the data is high-dimensional, SHAP-S will be more impacted by data sparsity, producing higher observation bias. 

\begin{figure}[tbh!]
\centering
    \begin{subfigure}[b]{0.48\textwidth}
         \centering
         \includegraphics[width=\textwidth]{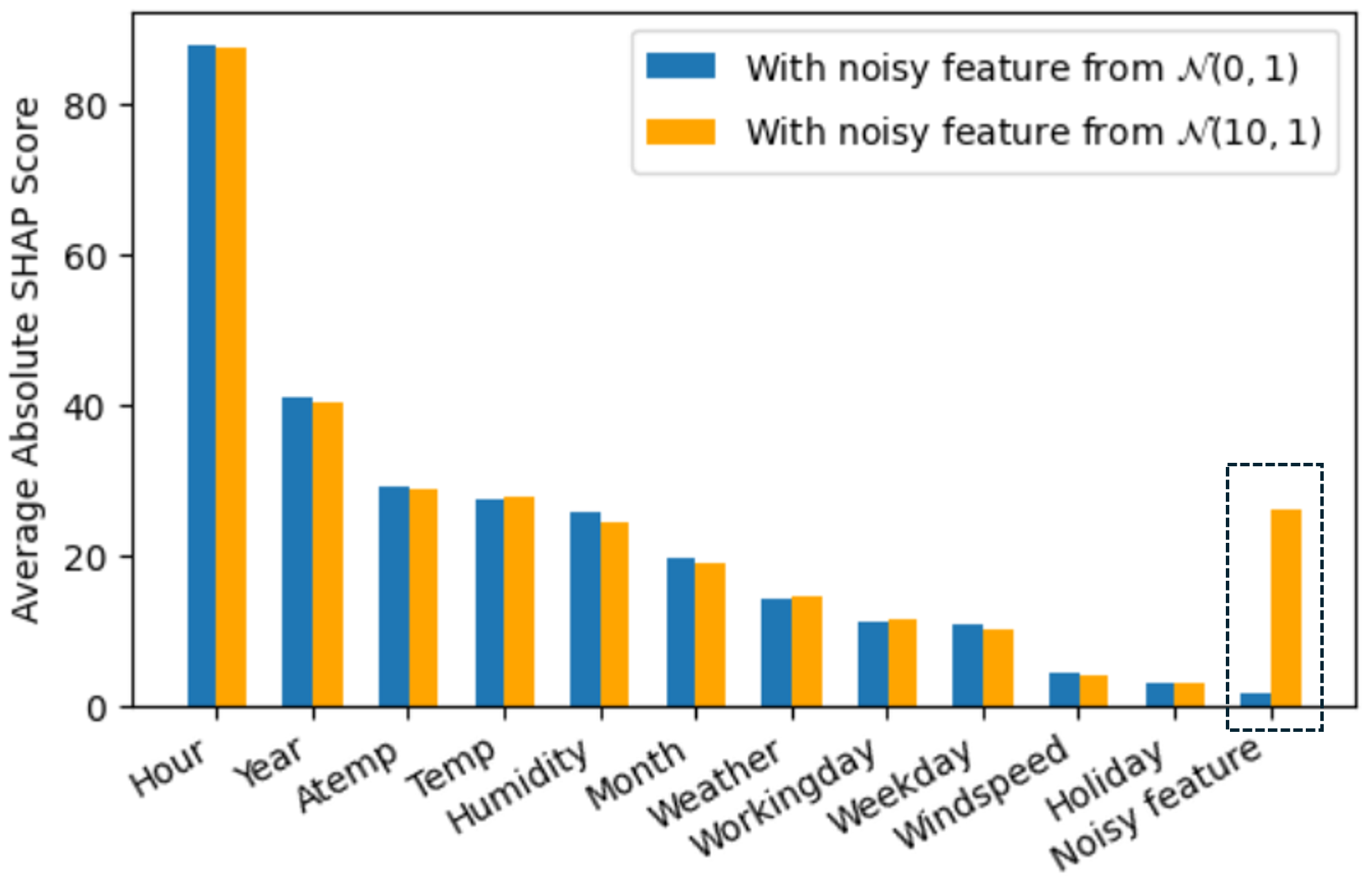}
         \caption{Bike Sharing Dataset}
         \label{fig:noisy_bike}
    \end{subfigure}
    \vspace{1mm}
    \begin{subfigure}[b]{0.48\textwidth}
         \centering
         \includegraphics[width=\textwidth]{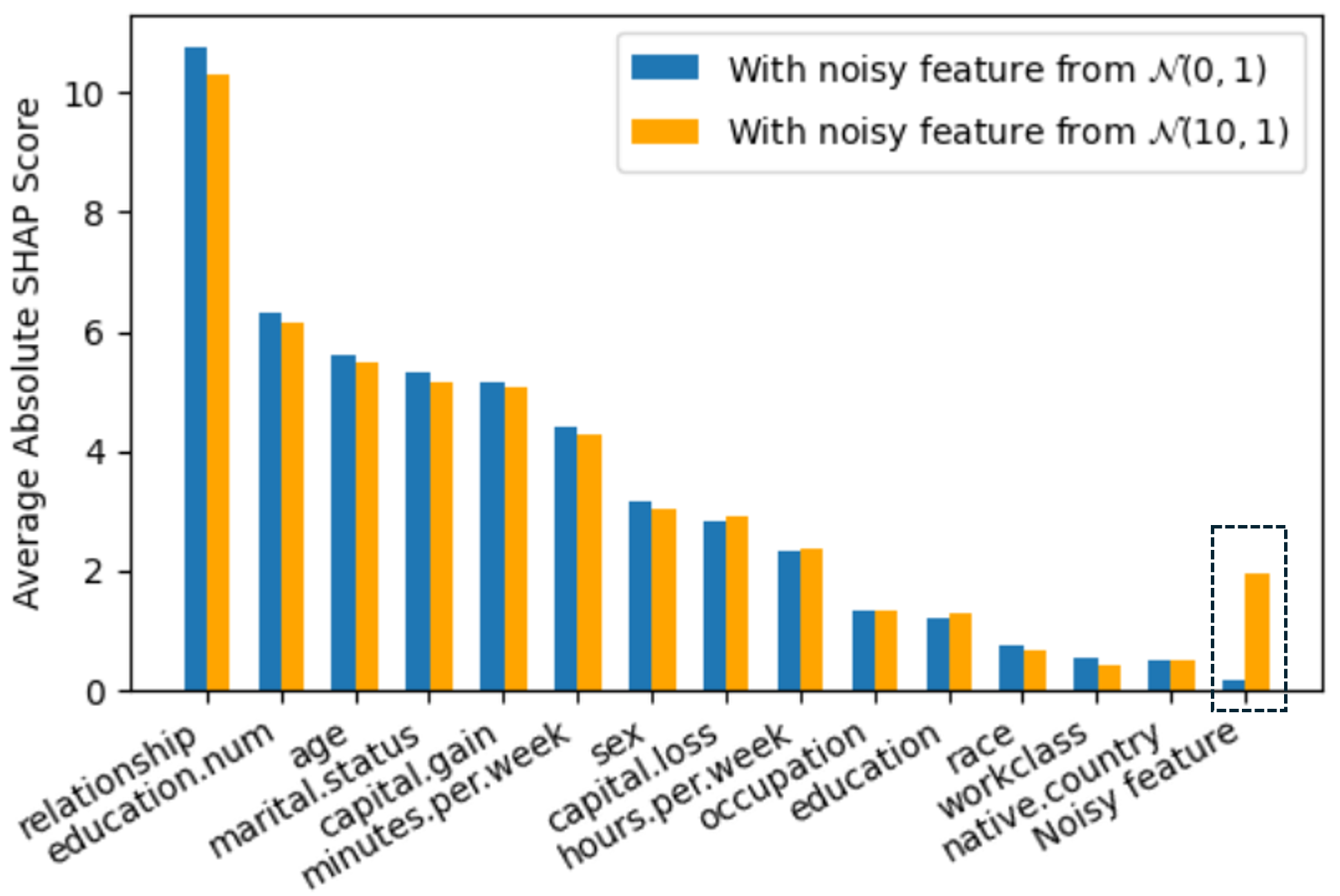}
         \caption{Census Income Dataset}
         \label{fig:noisy_income}
    \end{subfigure}
    
\caption{Average absolute feature attributions given by SHAP-S on 100 predictions where the noisy feature comes from either $\mathcal{N}(0,1)$ or $\mathcal{N}(10,1)$.}
\label{fig:exp4}
\end{figure}

As discussed in Section \ref{sec:error-data-smoothing}, even if the surrogate model has an overall good fit on a large explaining set, SHAP-S can still be over-informative on low-density regions where data sparsity persists. To verify this remark, we generate a noisy feature from a mixed Gaussian distribution: $Z \sim \mathcal{N}(0,1)$ with probability 0.999 and $Z\sim \mathcal{N}(10, 1)$ otherwise. For each dataset, we train a surrogate model on the whole training set with this noisy feature added. Even when the explaining set is large, the values from $\mathcal{N}(10,1)$ are still sparse, so the surrogate model is easy to overfit at points with $Z \sim \mathcal{N}(10,1)$. To see this, we use the SHAP-S feature attribution that utilizes the trained surrogate model to explain 100 predictions where $Z \sim \mathcal{N}(0,1)$ versus where $Z \sim \mathcal{N}(10,1)$. The feature attribution results are plotted in Figure \ref{fig:exp4}. We can see that, in both datasets, even with a surrogate model trained on a large explaining set, SHAP-S still assigns high importance to noisy features if given predictions with $Z \sim \mathcal{N}(10,1)$. This noisy feature should be given 0 importance because it is sampled independently from all other features. 

\section{Conclusions}

We proposed a unified error analysis framework for informative SVAs. Our framework stems from the estimation and approximation errors arising from estimating the conditional removal function. These errors correspond to observation and structural bias, which generate feature attributions that are respectively over- or under-informative. We apply our error analysis to discern potential errors in various existing SVA techniques. Carefully designed experimentation verifies our theoretical analysis. Future work can utilize our error analysis framework to develop new SVA methods that can effectively mitigate both under- and over-informativeness.

%
%

\begin{credits}
\subsubsection{\ackname} This study was funded by Mitacs (grant number IT33843), and Daesys Inc.

\end{credits}
%
%
%
\bibliographystyle{splncs04}
\bibliography{references}
%




\end{document}